%% file: Multi_Bird_Tracking_WACV2025.tex
\documentclass[10pt,twocolumn,letterpaper]{article}

\usepackage[pagenumbers]{wacv} 

\usepackage{graphicx}
\usepackage{amsmath}
\usepackage{amssymb}
\usepackage{booktabs}
\usepackage{array}
\usepackage{xcolor}
\usepackage{balance}

\usepackage{subcaption}

\usepackage[pagebackref,breaklinks,colorlinks]{hyperref}

\usepackage[capitalize]{cleveref}
\crefname{section}{Sec.}{Secs.}
\Crefname{section}{Section}{Sections}
\Crefname{table}{Table}{Tables}
\crefname{table}{Tab.}{Tabs.}
\newcommand{\gh}[1]{\textcolor{blue}{G.H: #1}}

\def\wacvPaperID{2872} 
\def\confName{WACV}
\def\confYear{2025}
\begin{document}
\title{Context-Aware Outlier Rejection for Robust Multi-View 3D Tracking of Similar Small Birds in An Outdoor Aviary}
\author{Keon Moradi,
Ethan Haque, Jasmeen Kaur,
Alexandra B. Bentz,
Eli S. Bridge, Golnaz Habibi\\
{\tt\small\{keon.moradi,ethan.a.haque-1,jkaur,abentz,ebridge,golnaz\}@ou.edu}
\\
\\The University of Oklahoma, OK, USA, 73019}
\maketitle
\begin{abstract}
This paper presents a novel approach for robust 3D tracking of multiple birds in an outdoor aviary using a multi-camera system. Our method addresses the challenges of visually similar birds and their rapid movements by leveraging environmental landmarks for enhanced feature matching and 3D reconstruction. In our approach, outliers are rejected based on their nearest landmark. This enables precise 3D-modeling and simultaneous tracking of multiple birds. By utilizing environmental context, our approach significantly improves the differentiation between visually similar birds, a key obstacle in existing tracking systems. Experimental results demonstrate the effectiveness of our method, showing a $20\%$ elimination of outliers in the 3D reconstruction process, with a $97\%$ accuracy in matching. This remarkable accuracy in 3D modeling translates to robust and reliable tracking of multiple birds, even in challenging outdoor conditions. Our work not only advances the field of computer vision but also provides a valuable tool for studying bird behavior and movement patterns in natural settings. We also provide a large annotated dataset of 80 birds residing in four enclosures for 20 hours of footage which provides a rich testbed for researchers in computer vision, ornithologists, and ecologists. Code and the link to the dataset is available at \href{https://github.com/airou-lab/3D\_Multi\_Bird\_Tracking}{https://github.com/airou-lab/3D\_Multi\_Bird\_Tracking}.
\end{abstract}
\input{introduction}
\input{related_work}
\input{dataset}
\input{method}

\input{result}

\input{conclusion}
\balance
{\small
\bibliographystyle{ieee_fullname}
\bibliography{Multi_Bird_Tracking_WACV2025}
}\end{document}

%% file: introduction.tex
\section{Introduction}
\label{sec:intro}
 Characterizing complex behaviors among animals living in naturalistic settings is crucial in biological and ecological sciences. Behavioral biologists have historically used continuous observations of one focal animal or scan sampling of groups~\cite{bate2021}. These approaches are labor-intensive and only capture a small subset of important social behaviors for a few individuals at a time. To record behavioral data at a finer scale, simultaneously across all individuals in a population, more advanced behavioral tracking technology is needed. Recent advances in computer vision and deep learning have made this a reality; however, to-date these techniques have largely been restricted to laboratory model species~\cite{kara2021} and mammals ~\cite{gosz2021,bala2020,marks2022}, while work in avian species is still limited~\cite{fang2021}. Birds are conspicuous, largely diurnal, and our knowledge of avian natural history and ecology is more extensive than that of any other vertebrates~\cite{koni1989}. Thus, birds could offer unique insights into complex social behaviors if we could appropriately track them in naturalistic settings. However, tracking birds in an outdoor environment is challenging due to the similarity of individual birds, their small size, their rapid movements, and potentially complex backgrounds. 
 
 Despite the advances of computer vision in the fields of advanced mobility and robotics, we lack extensive studies that apply advanced 3D computer vision to tracking animals consistently and reliably. Most current work is not generalizable to new environments; are limited to 2D detection and tracking; and often fail to track animals with rapid movement, such as flying birds. Multi-view technology has been used recently in the application of computer vision to provide accurate 3D construction for detection and tracking of objects in a 3D world. Multi-view tracking has also advanced robotics and autonomous driving by providing affordable sensing techniques compared to LiDAR technology. However, applying 3D computer vision technology to track highly dynamic animals such as birds is not trivial. Developing methods that can accurately capture and analyze complex flight behaviors and social interactions in settings that closely mimic natural environments remains challenging. In this paper, we focus on detecting and tracking visually similar birds, and we investigate and integrate combinations of techniques that optimize 3D-reconstruction results within a flight cage or aviary. 

\subsection{Main Contribution}
The main contributions of this paper are listed as follows:
\begin{itemize}  
    \item  Present a novel outlier rejection algorithm for feature matching which is based on the environmental context, called landmarks, to achieve an accurate 3D reconstruction for robust tracking.
     \item Propose a robust multi-view, multi-object tracking of visually similar birds in an aviary with a natural background. This leads to more consistent tracking, fewer ID switching, and fewer missing tracks, all of which improve the overall quality of tracking.
    \item Provide a large, annotated dataset of birds in an aviary for further study of animal behavior and spatio-temporal tracking.
\end{itemize}

%% file: related_work.tex
\section{Related Work}
\label{sec:related}
Tracking animals, especially birds that are visually similar to each other and their surroundings, produces several challenges such as occlusions and three-dimensional movements. However, in recent years, tracking animals with computer vision has emerged as a robust field of inquiry. Many studies have attempted to address these previously mentioned issues through multi-camera systems and computer vision techniques. This work is foundational to our approach to automated bird tracking. For instance,~\cite{xiao2023multi} presented a system for analyzing songbirds within a multi-view 3D aviary. Their work primarily addressed occlusions and variation in bird appearance in 3D space, using stereo matching and multi-view tracking. They also utilized a challenging dataset called WILD, which made their combination of detection software, like Mask R-CNN and Background Subtraction mask, more novel. This is also crucial in our segmentation technique. However, while they relied on stereo matching for 3D reconstruction, our approach relies more on the context of the birds, specifically its context-aware outlier rejection which improves the accuracy of feature matching and 3D reconstruction.

In recent years in advances in multi-view tracking, as outlined in \cite{teepe2024liftingmultiviewdetectiontracking}, there is a more involved priority and emphasis now on incorporating contextual information for enhanced feature matching and multi-object scenarios. Their work also use a landmark-based approach, although in a different domain, complements our methodology in leveraging the environment refine 3D reconstruction and reduce error; however, our method extends this concept by utilizing Voronoi diagrams for a more robust context-aware outlier rejection, especially in dynamic aviary settings. Similarly, \cite{wald2024} creates a system that essentially estimate the tracks of 3D-poses, called 3D-MuPPET, of similar animals such as pigeons, using multi-camera setups in controlled environments. While their work is focused on pose estimation in controlled conditions, our work focuses on more outdoor settings and environmental cues to handle harsh conditions and occlusions, aiming for enhanced tracking accuracy.

\cite{naik2023} attempted to address the issue of identifying the 3D positions of keypoints on birds in a similar aviary environment, though they indicate that their method of approximation works well for few, significant features only such as beaks and wings. We consider all keypoints in general obtained by the SIFT algorithm, and we do not consider any information relating to the quantity of features or the location of the features (\emph{i.e.,} keypoints) on the birds' bodies. A future work of interest would be in  comparing our results against their manually annotated dataset which includes ground truth values for the 3D trajectories of the birds throughout the intervals of their data. 

Furthermore, \cite{wu2010tracking} worked with bats, which are also visually similar and have rapid movements, to test which approach for tracking them was most successful in complex environments, either Reconstruction-then-Tracking (RT) or Tracking-then-Reconstruction (TR). Their findings show that the RT method performed better in datasets that have many occlusions which is another reason why we select the RT approach as our birds sometimes reside in highly occluded areas. However, we extend this by adopting a landmark-based matching system to be able to handle these occlusions as well as similarities.

Despite several studies in tracking visually similar animals, they are restricted to controlled lab environments such as~\cite{badger2020} and~\cite{gosztolai2021liftpose3d}. Our work involves highly dynamic outdoor setting including small birds, with natural lighting situations. By incorporating our landmark-based outlier rejection, we address the difficulties of matching visually similar birds, while also reducing ID switches and obtaining more consistent 3D tracking.

%% file: dataset.tex
\section{Dataset}
\label{sec:data}
 We have an aviary dataset which includes 20 hours of footage of 80 birds in an outdoor aviary. We utilize five GoPro cameras ($1920\times1080$ pixels, 30 FPS) aligned at various positions throughout an outdoor aviary enclosure (27.2\ \text{m}\textsuperscript{3}). The aviary is enclosed in hardware cloth and exposed to natural light cycles. Video recordings were synced using GoPro software to ensure proper temporal alignment. 
\begin{figure*}[ht!]
\centering
\begin{subfigure}{0.19\textwidth}
    \includegraphics[width=\textwidth]{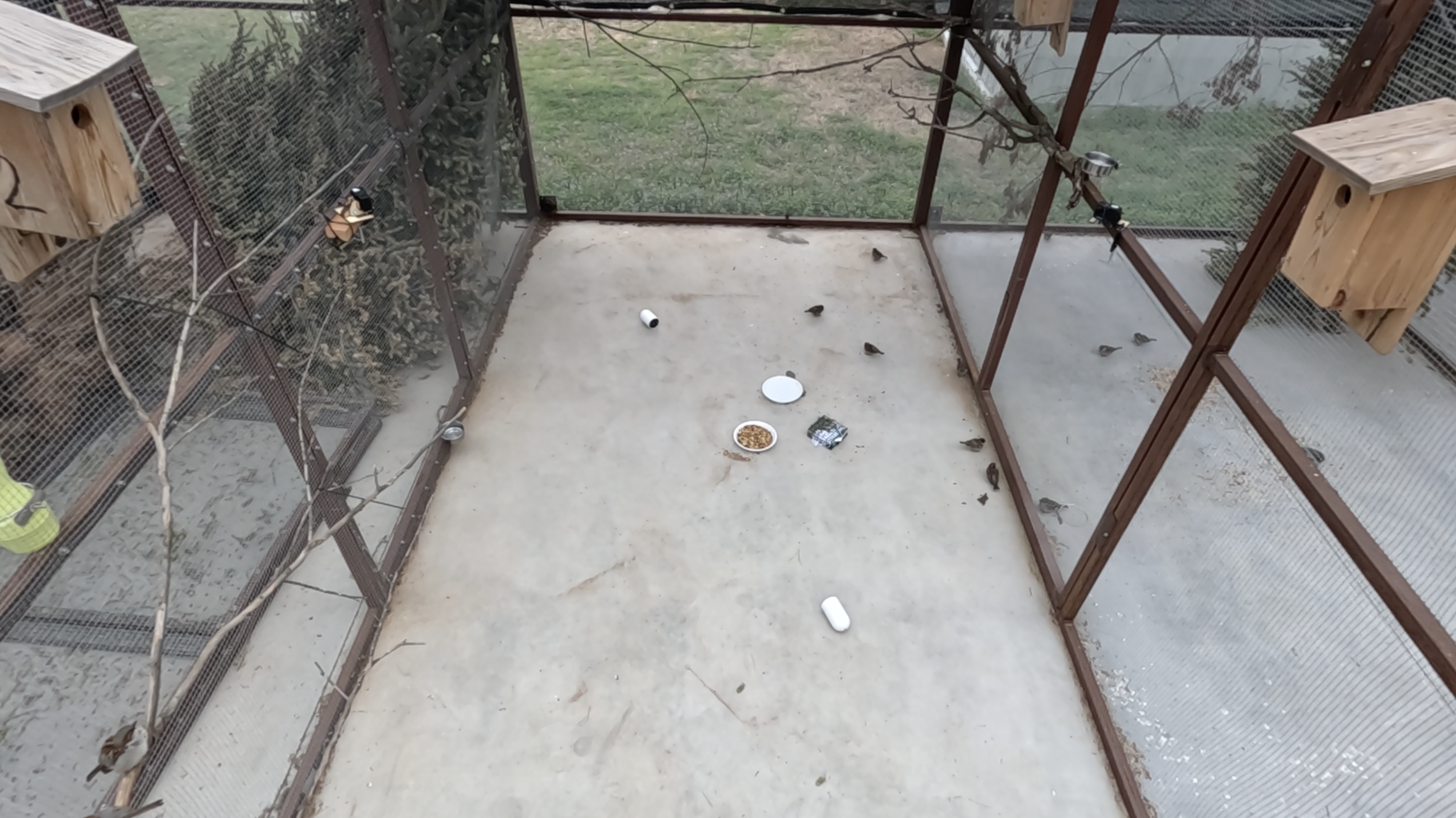}
    \caption{Camera view 1}
    \label{fig:camera1}
\end{subfigure}
\begin{subfigure}{0.19\textwidth}
    \includegraphics[width=\textwidth]{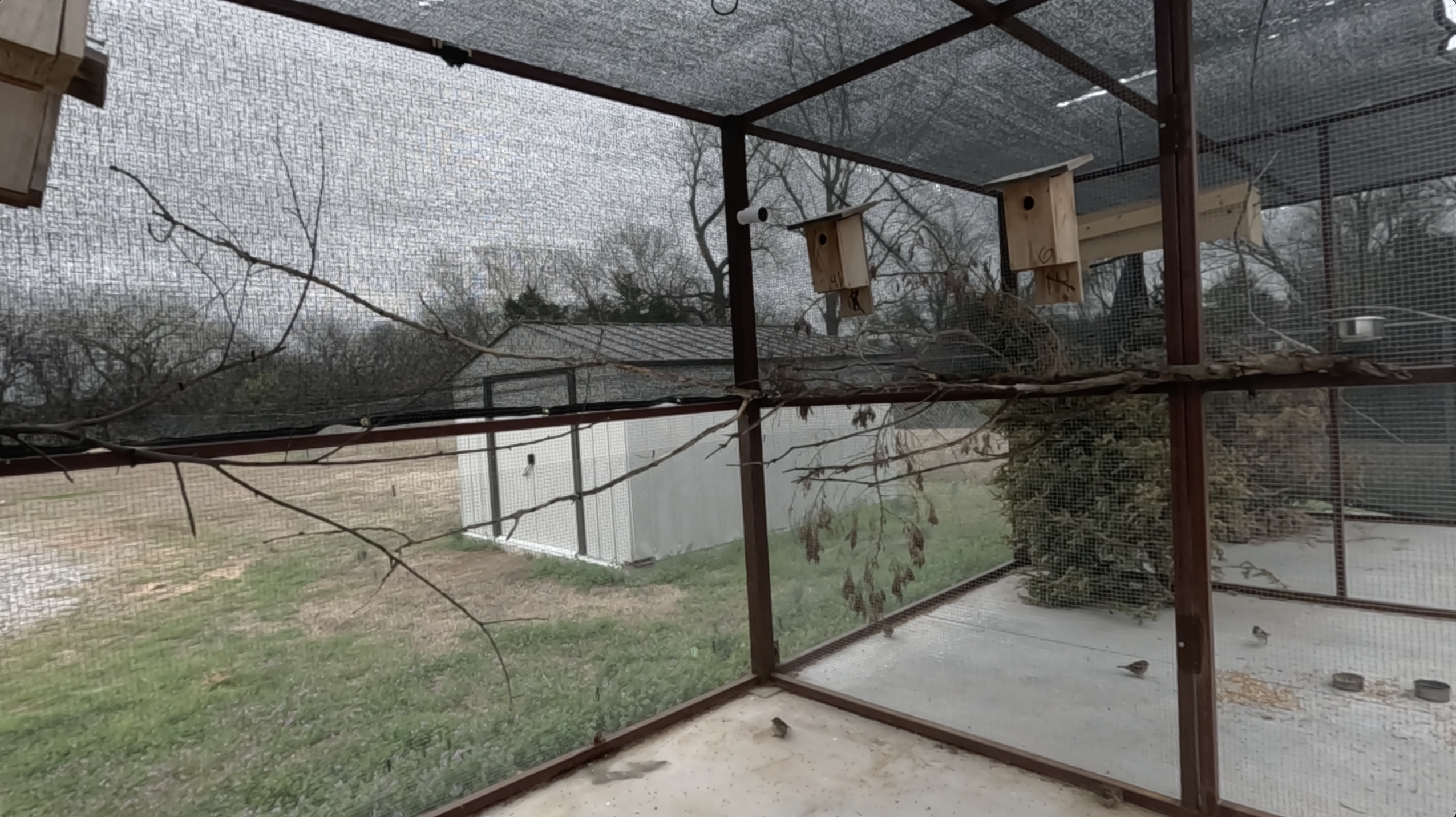}
    \caption{Camera view 2}
    \label{fig:camera2}
\end{subfigure}
\begin{subfigure}{0.19\textwidth}
    \includegraphics[width=\textwidth]{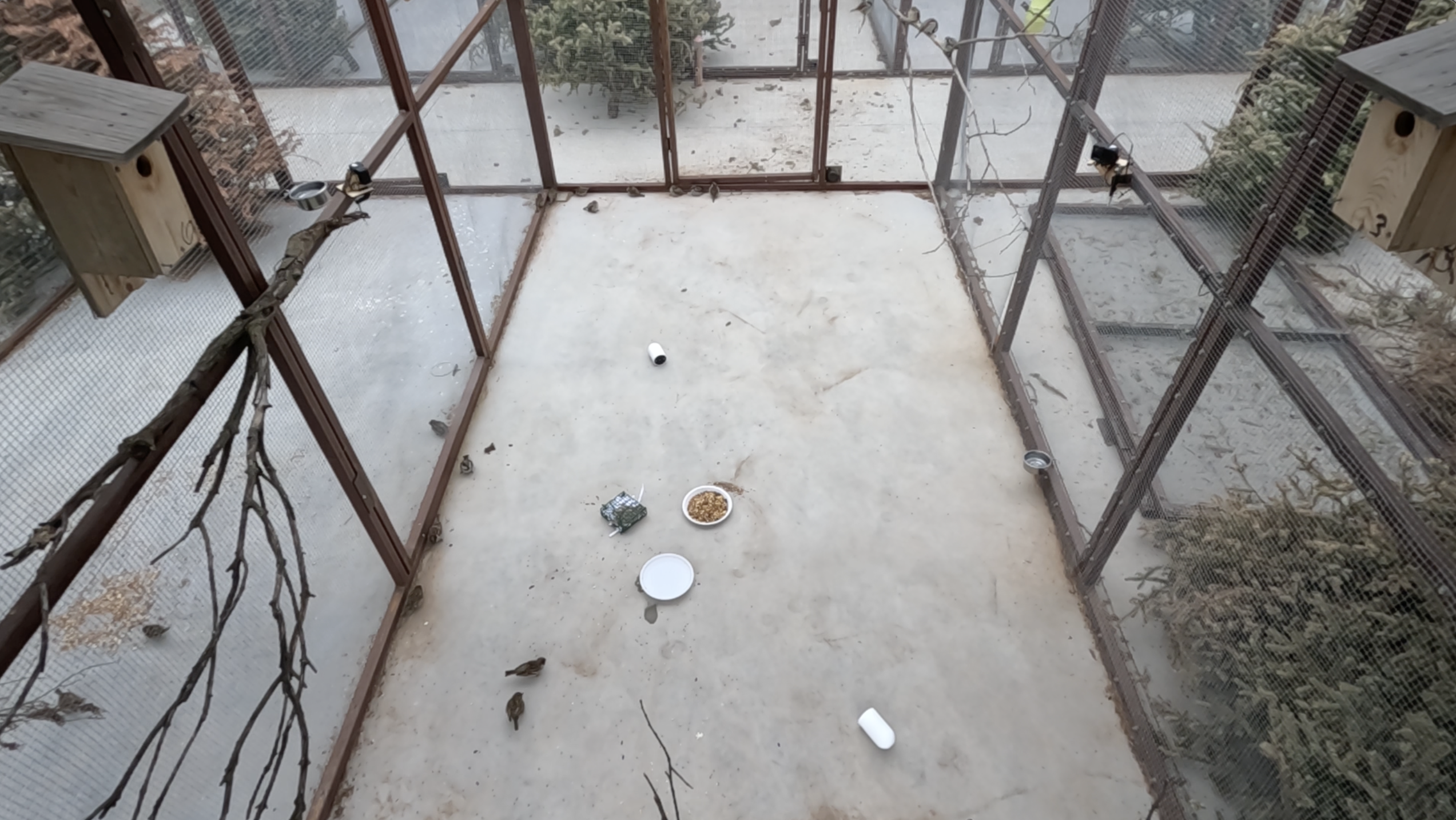}
    \caption{Camera view 3}
    \label{fig:camera3}
\end{subfigure}
\begin{subfigure}{0.19\textwidth}
    \includegraphics[width=\textwidth]{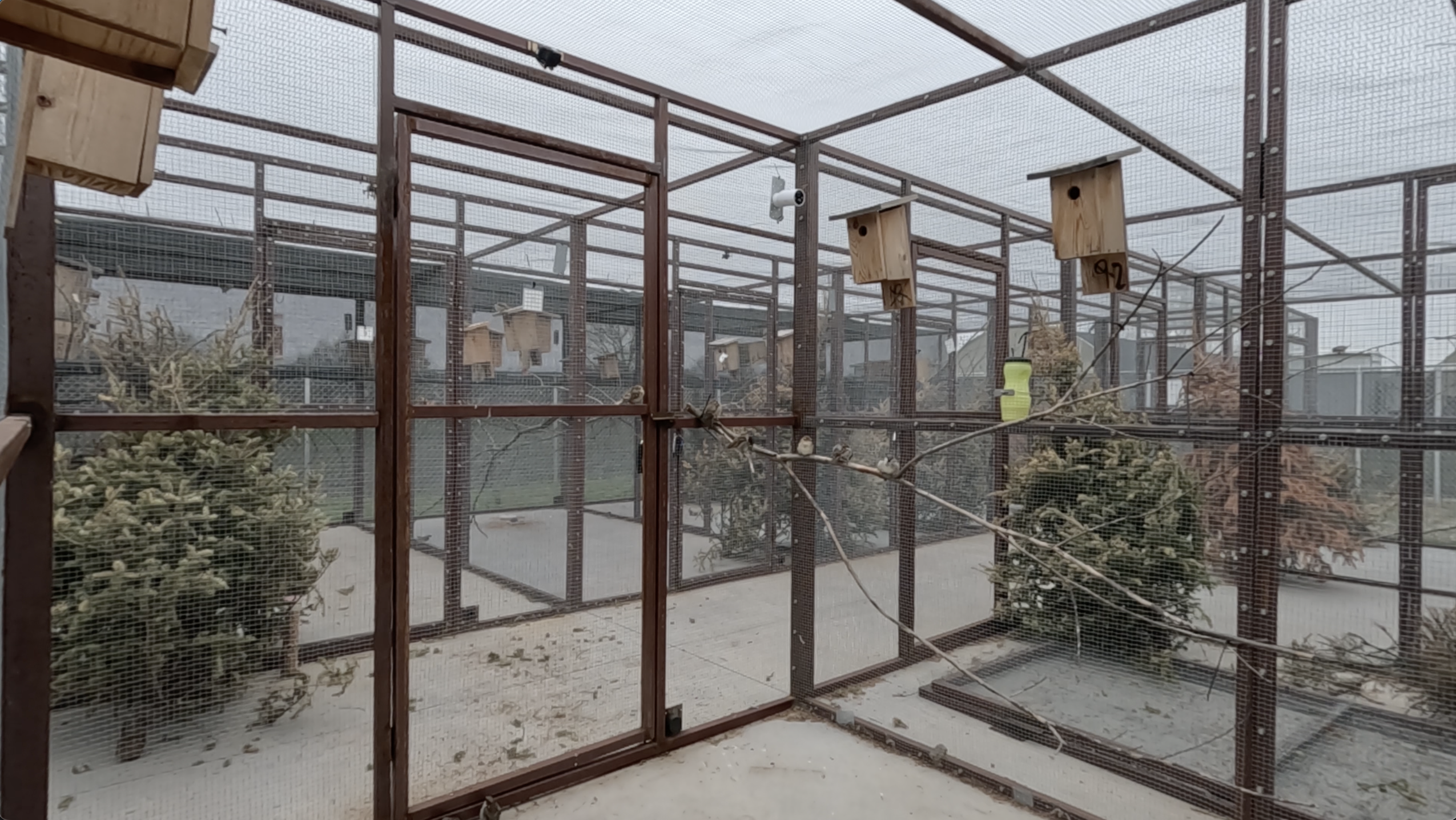}
    \caption{Camera view 4}
    \label{fig:camera4}
\end{subfigure}
\begin{subfigure}{0.19\textwidth}
    \includegraphics[width=\textwidth]{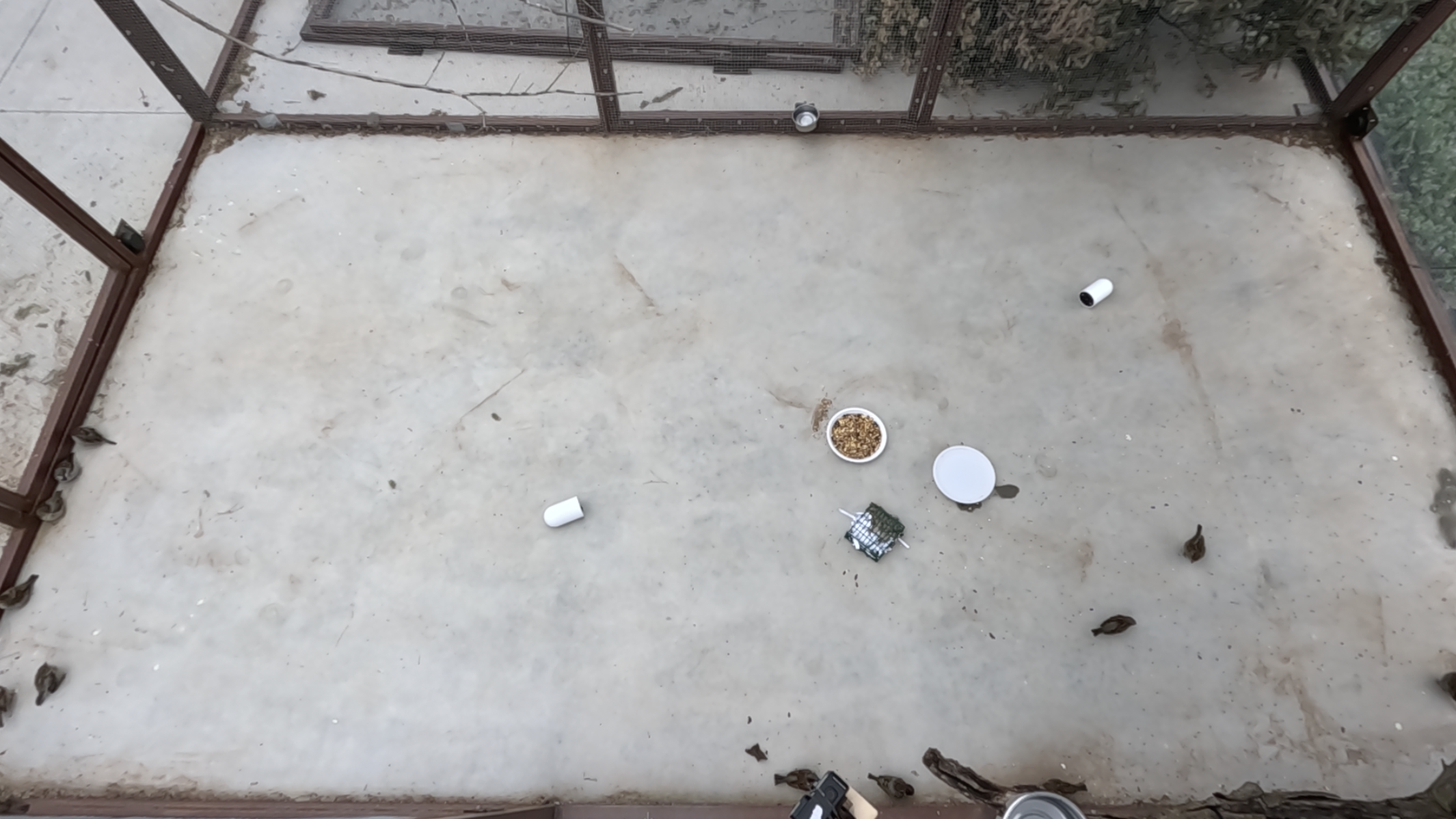}
    \caption{Camera view 5}
    \label{fig:camera5}
\end{subfigure}
\caption{Five camera views in one of the enclosures in the aviary, which is studied in this project.}
\label{fig:camera_views}
\end{figure*}
We housed four mixed-sex groups of 20 adult house sparrows (\textit{Passer domesticus}) in separate sections of the aviary. These 80 birds were captured locally and were given a unique combination of colored plastic leg bands for visual identification. Food and water were provided 
throughout the aviary in bowls, and wooden nest boxes were placed along the walls. Five corresponding camera views are studied in this paper and are shown in Figure~\ref{fig:camera_views}. 
The dataset is available at \href{https://osf.io/32zdt/}{https://osf.io/32zdt/}.

%% file: method.tex
\section{Multi-View Multi-Bird Tracking Overview}
\label{sec:overview}
\begin{figure*}
  \centering
  \includegraphics[width=\textwidth, trim={0 350 0 0}, clip]{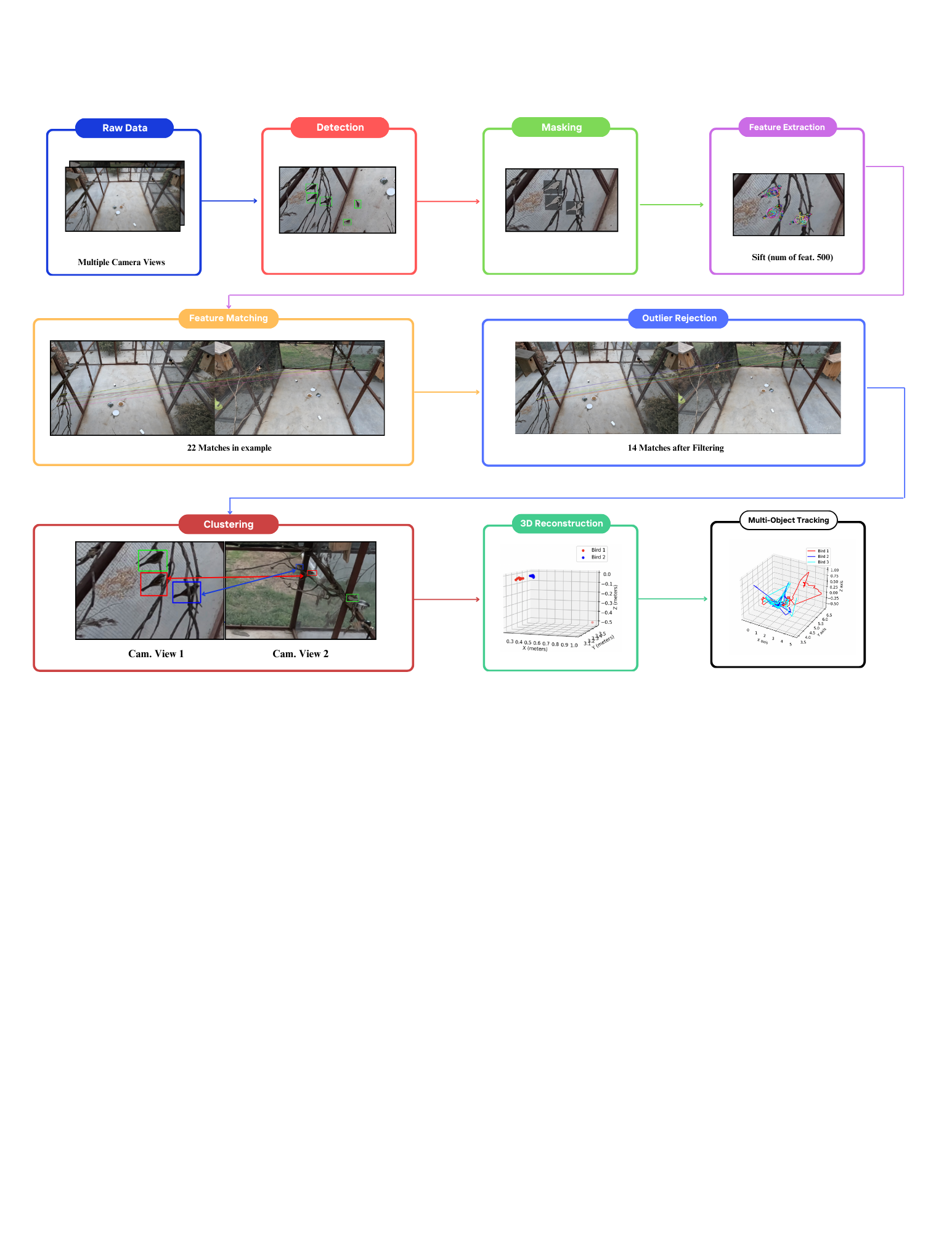}
  \caption{Proposed Workflow diagram: our multi-view 3D multi-bird tracking \label{fig:workflow}}
\end{figure*}

The workflow of this project begins operation on five synced videos from a single enclosure. Figure~\ref{fig:workflow} depicts the pipeline of our workflow  for 3D tracking of birds in an aviary using multiple cameras. As illustrated, the proposed algorithm has eight main steps: \textbf{(1) Object detection:}
 We first render each video as a sequence of individual images (frames), and we manually annotated the first 120 frames with boxed regions that indicated bird locations. These annotated frames were then used for training the YoloV5 model ~\cite{redmon2016}, which obtained bird detections for the remaining frames, resulting in .csv files that listed the bounded box coordinates of bird detections and other relevant data; \textbf{(2) Masking:} We applied an edge detection algorithm and laterally fill in pixels within bird detection bounded boxes to obtain a binary mask depicting bird detection and background features for each frame; \textbf{(3) Keypoint Extraction:} keypoints and corresponding descriptors are extracted only from "on" pixel regions of the mask using the SIFT keypoint extraction algorithm~\cite{lowe2004}. This ensures features are only extracted from the bird or their neighborhood. \textbf{(4) Feature Matching:} the extracted keypoints are used to match birds between camera views to help globally track similar birds across all camera views using the Brute-Force matching algorithm; \textbf{(5) Outlier Rejection:} we integrate our context-based outlier rejection method by constructing a Voronoi diagram using landmarks selected throughout our camera views as Voronoi coordinates constructing the Voronoi diagram, which is then layered on top of our image. We then use these 5 initialized Voronoi diagram objects (one per camera) to associate each bird to their nearest landmarks in  each video frame. Then, of the initial features/keypoints matched earlier, only the ones that also have matching nearest landmarks (landmark closest to their corresponding bird detection) are validated and kept; \textbf{(6) Clustering:} by keeping the bounding box index (\emph{i.e.,} bird) corresponding to each matching feature, the matched features are clustered; \textbf{(7) 3D reconstruction:} following the previous step, a 3D reconstruction representing the global view of our aviary is obtained using multi-view geometry and triangulation; \textbf{(8) Multi-object tracking:} each bird is tracked across five camera views by following its center in 3D coordinates. We utilize the tracking-by-detection technique, and we apply a Kalman filter to predict each bird's 3D position in the next frame. 
\section{Object Detection}
\label{sec:detection}
Previous work~\cite{xiao2023multi} used Mask R-CNN \cite{he2017mask} due to its ability to generate high-quality segmentation masks alongside object detection. However, considering the complexity and difficulties of our hardware and the large volume of data we need to process, Mask R-CNN proved to be less efficient. Instead, we use YOLOv5, which offers a more streamlined and faster approach to object detection while maintaining robust accuracy. The decision to switch to YOLOv5 was made due to its real-time processing capabilities and the need to handle high frame rates without sacrificing detection quality. YOLOv5's backbone architecture, coupled with its efficient detection head, allowed us to process each frame more rapidly, which was essential given our dataset's size and the dynamic nature of the aviary environment. Furthermore, as in Table~\ref{tab:detection_results}, YOLOv5 base model detects more birds compared to Mask-RCNN. This led us to consider YOLOv5 as object detection backbone.

Initially, when running YOLOv5 on our dataset, it detected approximately 1448 birds in the first video which is only $16\%$ of the birds. Recognizing the need for enhanced detection accuracy, especially given the challenging conditions of a natural aviary setup (e.g., occlusions, similar appearances of birds), we undertook a fine-tuning process. This involved augmenting the model with an additional 600 newly annotated frames specifically chosen to capture a variety of challenging scenarios.

The fine-tuning significantly improved the model's performance, with detections increasing to over 6000 detection in the same video—a nearly 7.5-fold improvement. This dramatic increase in detection counts proved to be fruitful, in addition to adjusting hyper-parameters like learning rate and batch size. Continuous fine-tuning was also employed as new data became available, allowing the model to adapt to different lighting conditions, angles, and bird behaviors. This iterative process ensured that the model remained robust and could generalize well across different frames and scenarios.

As shown in Table~\ref{tab:detection_results}, the trend of improvement across all of the different models and fine-tuning was done to receive the best possible results within the detection step before proceeding through our pipeline. We took the same video sample of Figure~\ref{fig:camera_views} and ran it through only 1 minute at 30FPS, or in total 1800 frames for each model, and we documented how many bird labels were predicted. Note that Mask RCNN and YOLOv5 Base models detected object classes other than birds. The numbers in Table~\ref{tab:detection_results} reflect the estimation of the number of birds detected only.  The fine-tuning gave significant improvement to the detection step, especially the first 300 annotated frames that we trained on, and returned a closer estimate of detected bird labels after an additional 300 frames were annotated, totaling 600 annotated frames, all 600 frames are a combination of all 5 camera angles, upon which our YOLOv5 model was trained on. Note, we only used one YoloV5 Model, the base pre-trained, and continue to further fine-tune the same model after each time more time is collected with all camera angles included at the same time.

\begin{table}
\centering
\resizebox{\columnwidth}{!}{%
\begin{tabular}{|l|c|}
\hline
\textbf{Model/Method} & \textbf{Detected Bird Labels} \\
\hline
Mask RCNN (Base Model) & 823 \\
YOLOv5 (Base Model) & 1448 \\
Fine-Tuned YOLOv5 (300 Annotated Frames) & 5548 \\
Further Fine-Tuned YOLOv5 (600 Annotated Frames) & 6308 \\
\hline
Estimate of Ground Truth (4-5 Birds, 1800 Frames) & 7200 - 9000 \\
\hline
\end{tabular}%
}
\caption{Result of bird detection by running different backbones.}
\label{tab:detection_results}  
\end{table}

\section{Feature Extraction and Matching}
\label{sec:feature}
\begin{figure*}[ht]
    \centering
    \begin{subfigure}[]{0.5\textwidth}
        \centering   \includegraphics[width=0.85\textwidth, trim={0 0 0 0},clip]{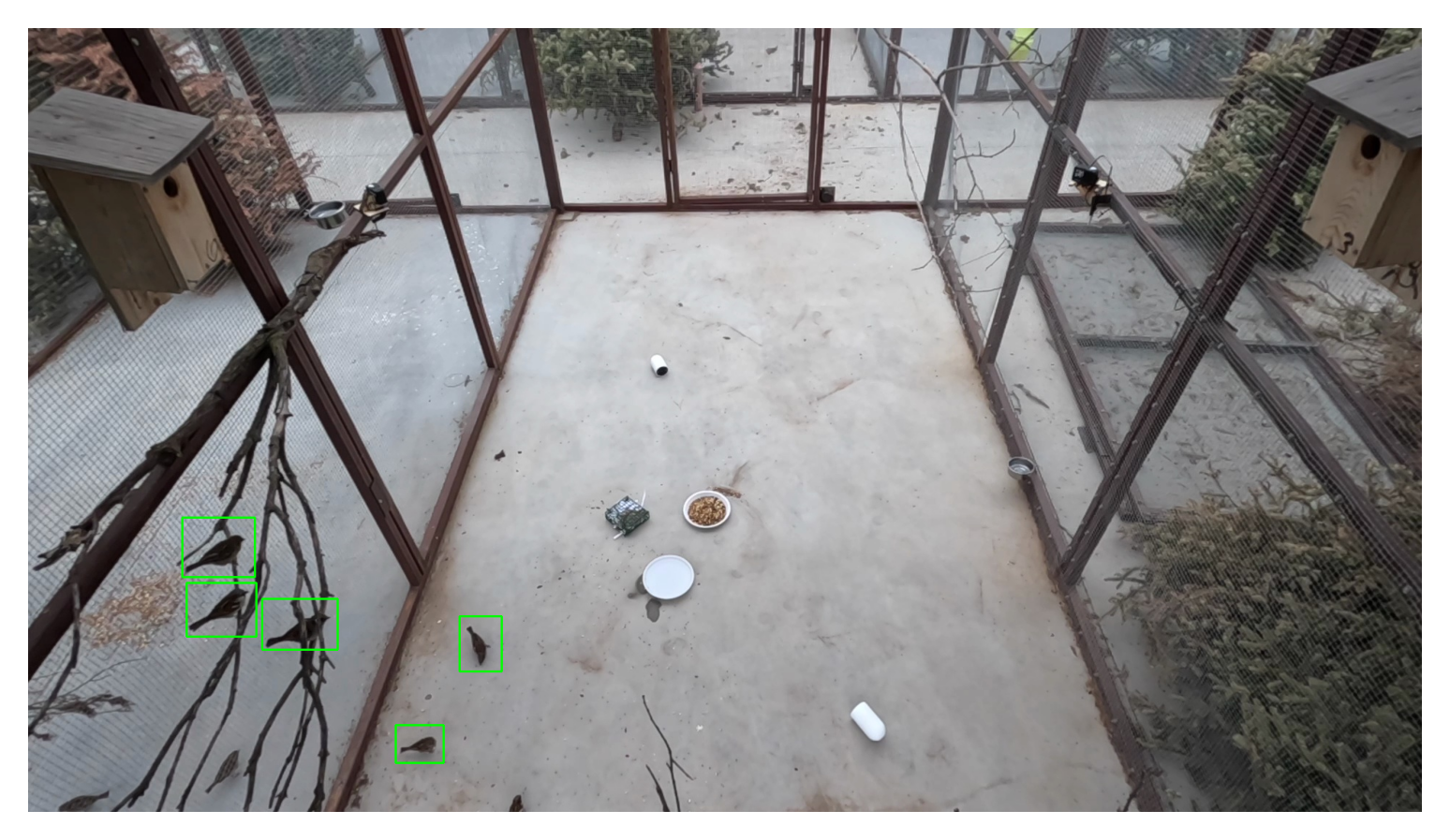}
    \end{subfigure}%
    \begin{subfigure}[]{0.5\textwidth}
        \centering
        \includegraphics[width=0.85\textwidth, trim={0 0 0 0},clip]{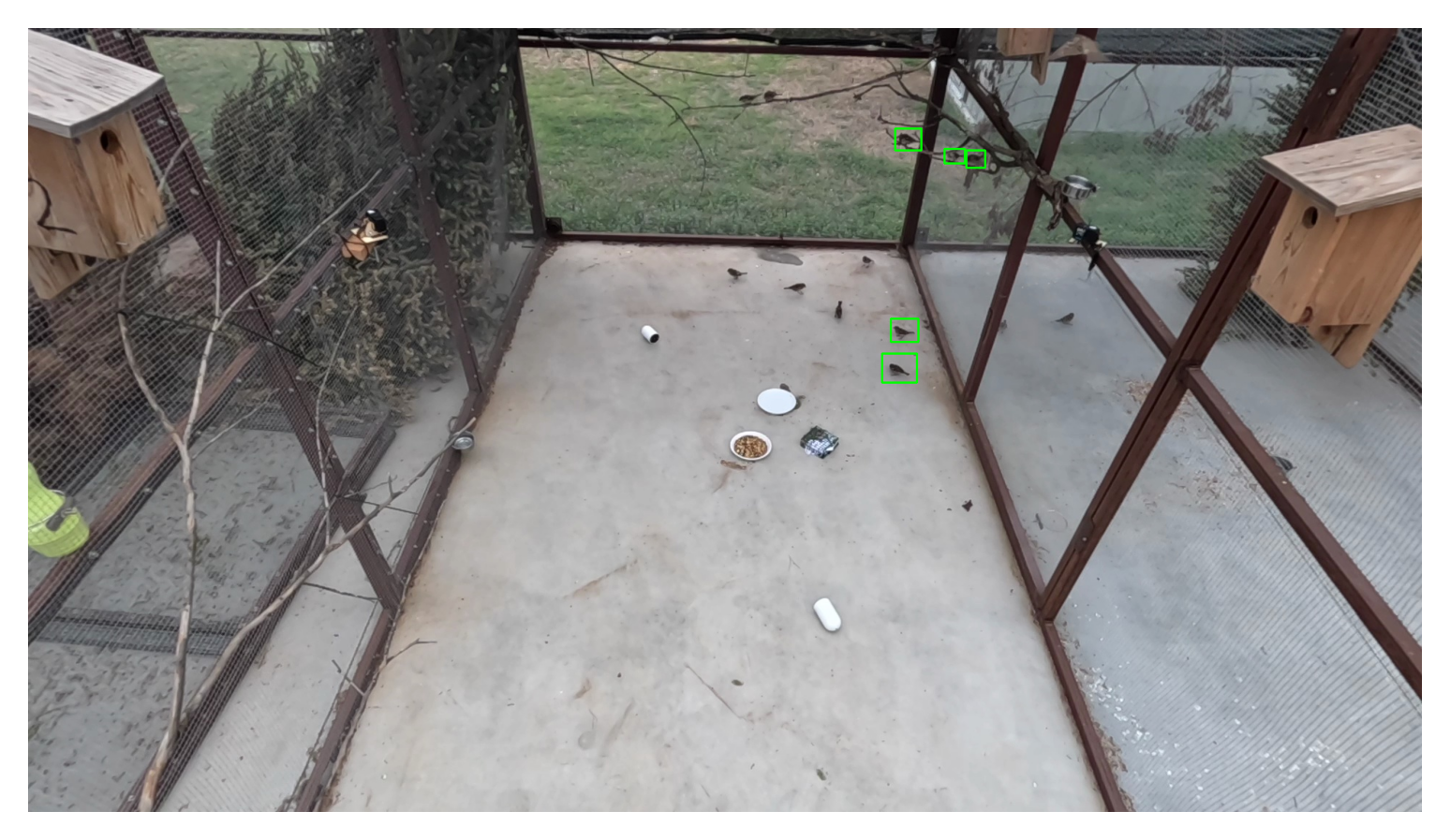}
    \end{subfigure}


    \begin{subfigure}[]{0.5\textwidth}
        \centering
        \includegraphics[width=0.85\textwidth, trim={0 0 0 0},clip]{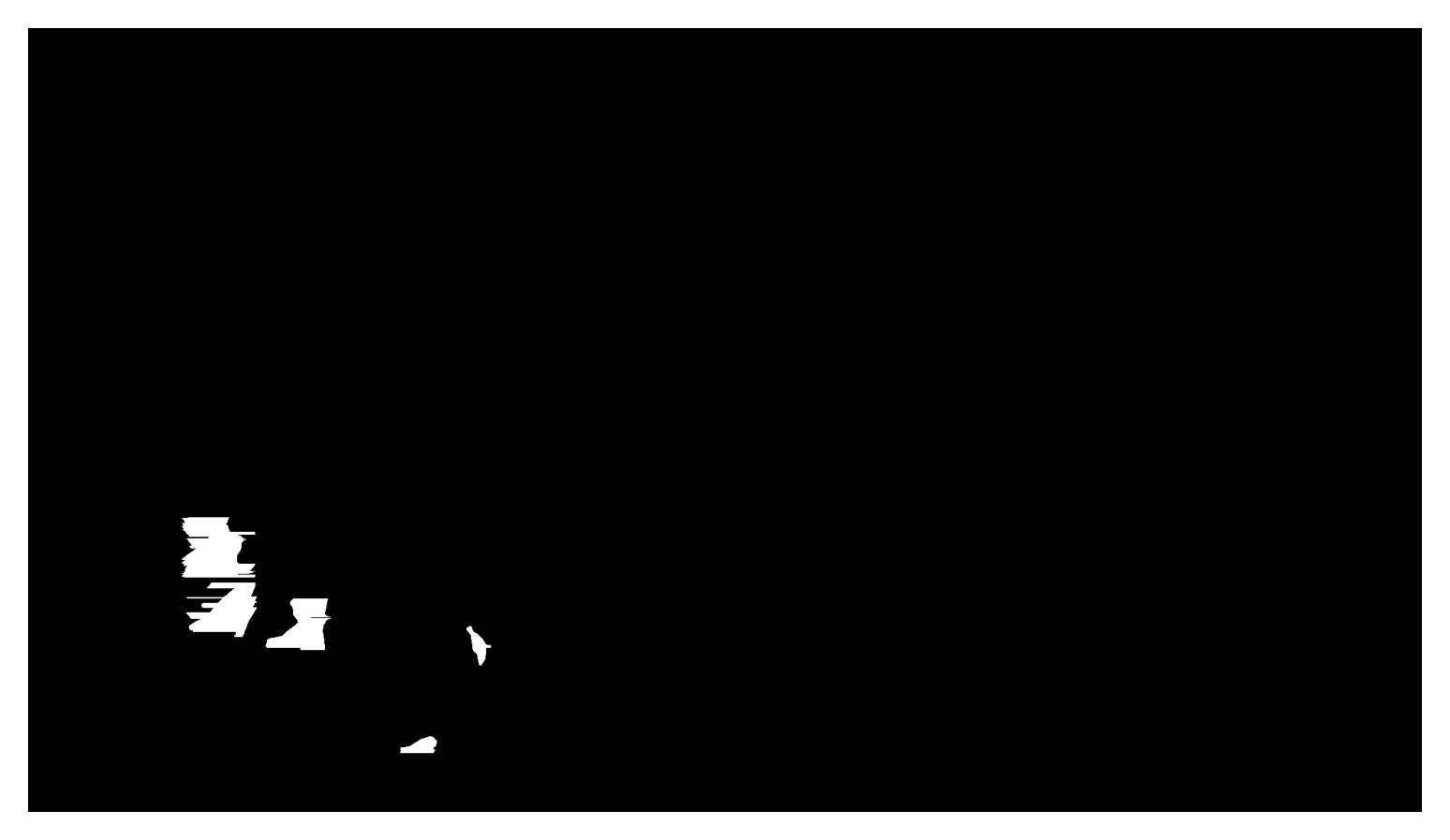}
        \caption{camera view 1}

    \end{subfigure}%
    \begin{subfigure}[]{0.5\textwidth}
        \centering
        \includegraphics[width=0.85\textwidth, trim={0 0 0 0},clip]{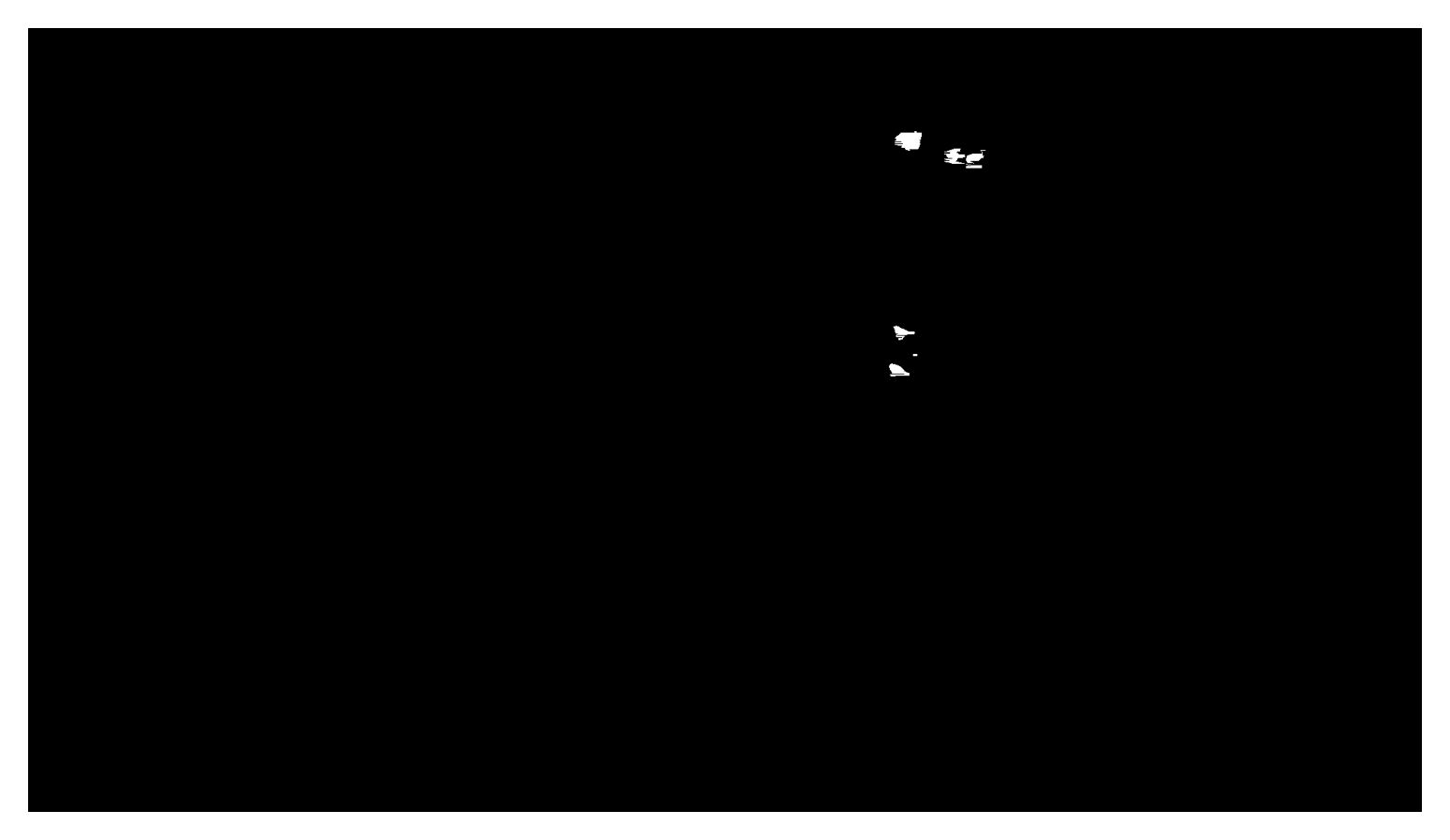}
        \caption{camera view 2}
    \end{subfigure}
   
    \caption{Masking Step. We consider masking algorithms, with the result shown in the bottom row.}
    \label{fig:camera13}
\end{figure*}

Image frames containing bird detections from the trained and tuned YOLOv5 model undergo our binary masking segmentation to obtain pixel regions that either directly or contextually indicate bird activity/relevant bird features. Our masking technique operates on image frames that contain bird detections by initially applying Canny edge detection on bounded boxes corresponding to bird detections~\cite{canny1986}. From this, we obtain a mask of our detections by laterally filling in the pixels between all pixels that are detected to be an edge within the pixel coordinate row of the frame. Figure~\ref{fig:camera13} shows the result of masking in two camera views.
Feature extraction is performed on masked image frames using the Scale-Invariant Feature Transform (SIFT)~\cite{lowe2004} to identify suitable interest points in the image that are appear invariant to scale and orientation.

\begin{figure}[!ht]
    \centering
    \begin{subfigure}[]{0.5\textwidth}
        \centering
        \includegraphics[width=0.9\textwidth, trim={0 0 0 0},clip]{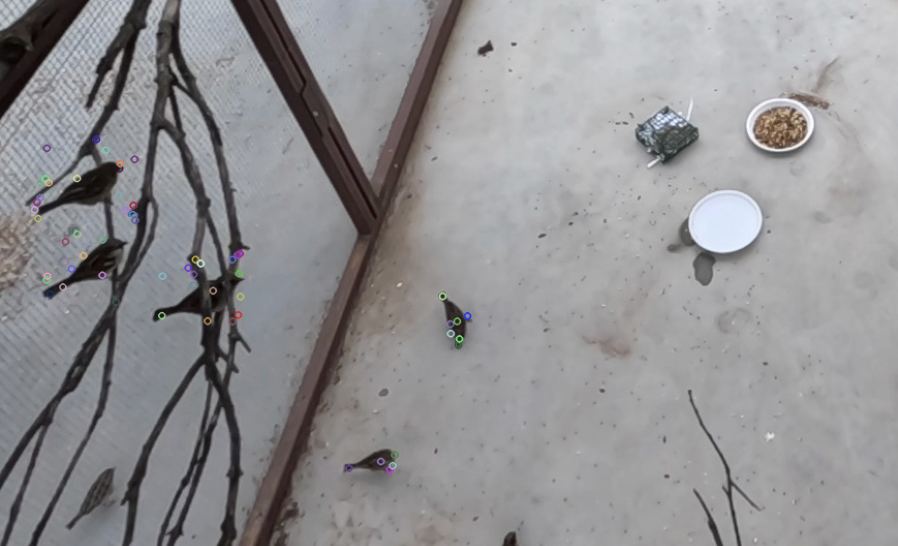}
    \end{subfigure}
    \begin{subfigure}[]{0.5\textwidth}
        \centering
    \includegraphics[width=0.9\textwidth, trim={0 0 0 0},clip]{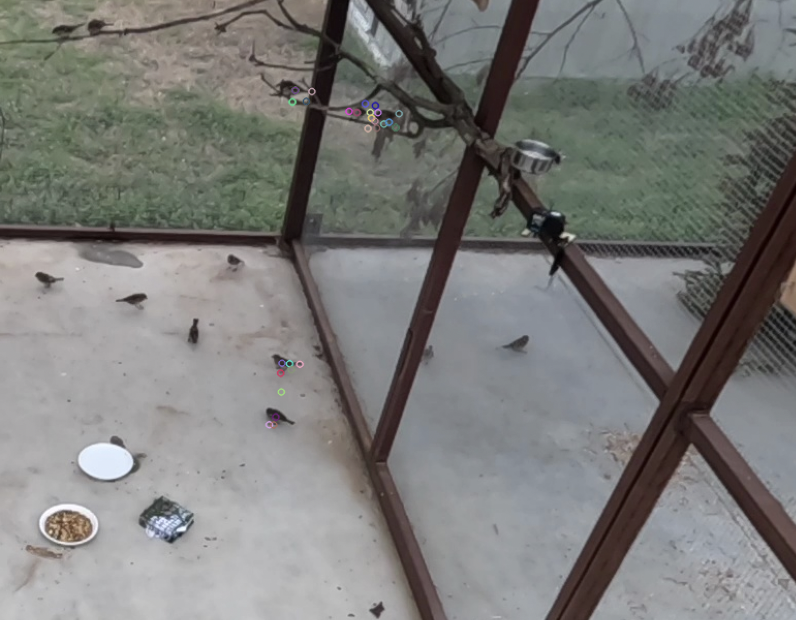}
    \end{subfigure}
    \caption{(a) Keypoints in the first camera view; (b) Keypoints in the second camera view}
    \label{fig:image3_with_keypoints}
\end{figure}
After obtaining a complete list of keypoints aligning with the bird detections in two camera views, we use Brute-Force feature matching combined with \textit{k}-Nearest Neighbors algorithm to find correspondences between keypoints in images. Figure~\ref{fig:image3_with_keypoints} shows an example of feature extraction inside the birds.
\section{Landmark based Outlier rejection}
\label{sec:outlier}
 Matching extracted features between birds at outdoor aviary has its own challenges which mainly lies in the physical similarities of each bird. Even for humans, the task of quickly identifying and matching birds in the aviary between camera views remains difficult due to the birds' dark color, similar shape, and frequently occlusive behavior. Though we are able to ensure that all extracted keypoints lie on birds as they are limited to the area of detected bounding boxes coordinates, all birds have a similar shape and color, especially in farther distance, which limits the accuracy of keypoint matches using standard approaches such as Brute Force Matcher algorithm. \cite{jaku2018} To improve matching accuracy and address this limitation, we propose a new approach that leverages the location of known landmarks to remove some outliers (\emph{i.e.,} incorrect matches). Our approach is based on Voronoi diagrams of the landmarks, such that two features are matched if they belong to the same voronoi cell.

\begin{figure}[!ht]
  \centering
\includegraphics[width=0.48\textwidth, trim={100 50 100 50},clip]{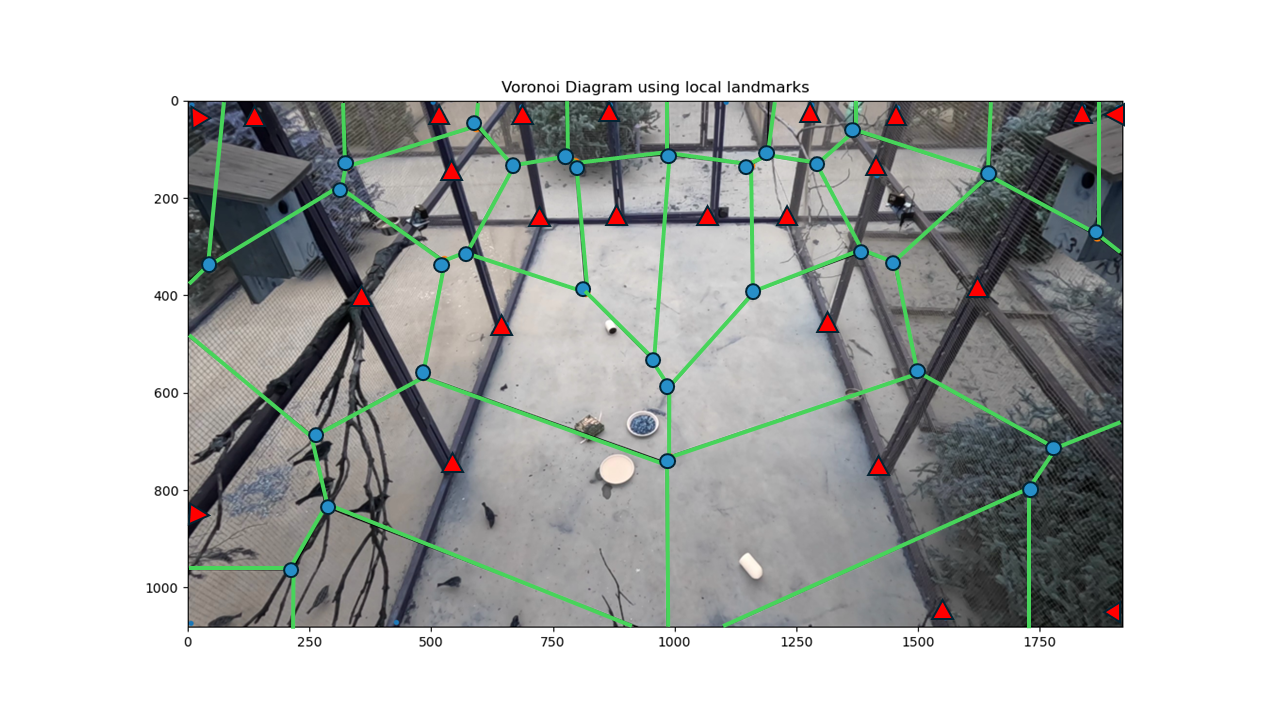}
  \caption{Locally constructed Voronoi diagram.}
  \label{fig:VoronoiExample}
\end{figure}

By definition, the Voronoi diagram of a set of points $P = \{p_{1}, ..., p_{n}\}$ in $\mathbb{R}^{d}$ creates a partition of $\mathbb{R}^{d}$ into $n$ regions, where all points in a region share a common closest point in the set $P$ according a distance metric $D(.,.)$~\cite{shah2009}. In other words, the region corresponding to a point $p \in P$ also contains all other points $q \in \mathbb{R}^{d}$ for which the following condition holds:
\begin{equation}
    \forall p' \in P, p' \neq p, D(q,p) \leq D(q,p').
\label{eq:voronoi_definition}
\end{equation}
By adapting the Voronoi diagram to be constructed from the local pixel coordinates of global landmarks across the aviary enclosure, we introduce a novel context-based outlier rejection algorithm for the previously obtained feature matches. We layer our initially constructed Voronoi diagram over each image frame, thus segmenting our image into regions uniquely identifiable by their local corresponding landmark. Figure~\ref{fig:VoronoiExample} depicts this segmentation technique, where the red triangles correspond to the locations of landmarks selected in the camera view. The solid green lines are Voronoi edges which create Voronoi cells with blue circles as Voronoi vertices.

Incorporating arbitrary distance metrics yields different variations of Voronoi diagrams. We assume Euclidean distance in our approach.
The Euclidean distance function measures the distance between two points $p, q \in \mathbf{R}^2$, $p = (p_1, p_2)$ and $q = (q_1, q_2)$ as such:

\begin{equation}
    d(p, q) = \sqrt{(p_1 - q_1)^2 + (p_2 - q_2)^2}
\label{eq:euclidean_distance}
\end{equation}

We calculate the Euclidean distance between each matched SIFT keypoint's location and all local landmarks in the camera view to identify the nearest landmark. For a match to be valid, the nearest landmarks for the same keypoint, as observed by both paired cameras, must agree. If the nearest landmarks differ, we reject the match, using context-based inference to eliminate mismatched birds.
\subsection{Voronoi cells with bounded edges}
To best of our knowledge, this is the first time that Voronoi diagram is used for image tessellation. Image segmentation using Voronoi diagram requires Voronoi cells with bounded edges. Otherwise, we cannot implement it in practice.  We claim this as part of the novelty of our work, as we derived a process to subvert the creation of "infinite edges". We first pad the image with the size of image diameter. Then, we add a set of "virtual landmarks" to the padded region, ensuring the bounded Voronoi diagram with closed boundaries spans the entirety of our image frame. 

\section{3D multi-object tracking of the birds}
\label{sec:track}
Our 3D-Tracking involves a multi-step process of calibrating cameras then reconstructing 3D positions, and finally tracking the birds in our multi-view setup. We achieved the calibration through manually estimating the intrinsic and extrinsic parameters using known dimensions and then refining with re-projection error minimizing. Once detections are matched camera views, 3D positions are made through triangulations, which goes toward the initial position for a Kalman filter for a smooth and accurate tracking over time and refined predictions based on new observations.

\subsection{Camera Calibration}
The process begins with the calibration of the cameras to determine the intrinsic and extrinsic parameters required for 3D Pose Estimation. We note that we needed to obtain the calibration's manually as this was not an automated process, and we used multiple methods such as OpenCV's Chessboard method \cite{bradski2008learning} to match similar points and objects of known fixed positions across pairs of camera angles in order to find the calibration. In addition, we utilized the concept of finding six known points of measurements that fit a pattern to also achieve a more accurate calibration. This process involved making note of the aviary's dimensions to get accurate measurements of the corners of the metal perimeter bars. For each camera, the intrinsic matrix \( K \), distortion coefficients \( D \), rotation vectors \( rvecs \), and translation vectors \( tvecs \) are used to compute the projection matrices. The projection matrix \( P \) for each camera is given by:

\begin{equation}
P = K [R | t]
\end{equation}

where \( R \) is the rotation matrix obtained from \( rvecs \), and \( t \) is the translation vector \( tvecs \). The projection matrices allow us to map 2D image coordinates to 3D space. After our initial calibration, we ran a verification between the corner points that we selected as our known points re-projected on to the image and ran an error check, and then manually further adjusted the parameters to minimize the error.

\subsection{3D reconstruction from multiple camera views}
Once the birds have been detected and matched across camera views, using the landmark-based feature matching described earlier, we proceed with the 3D-reconstruction of their positions. For each matched pair of detections across two camera views, the 3D position of the bird is calculated using triangulation. This process involves finding the intersection point of the lines of sight from the two cameras, which is an essential part of many computer vision applications\cite{DBLP:journals/corr/abs-1907-11917}. It essentially contains the essence of the methods in \cite{xiao2023multi} in regards to their reconstruction; however, our matching is different; however, the triangulation is essentially similar. Given the matched 2D coordinates \((x_1, y_1)\) and \((x_2, y_2)\) from the two views, the 3D point \( \mathbf{X} \) is obtained by solving the following equation:

\begin{equation}
\mathbf{X} = \text{triangulate}(\mathbf{x}_1, \mathbf{x}_2, P_1, P_2)
\end{equation}

where \( \mathbf{x}_1 \) and \( \mathbf{x}_2 \) are the homogeneous coordinates of the matched points, and \( P_1 \) and \( P_2 \) are the projection matrices for the two cameras~\cite{hartley2003multiple}.

\subsection{Multi-bird Tracking using Kalman Filter}
To ensure consistency and accuracy in tracking over time, we employ a Kalman filter with velocity tracking for each bird. The Kalman filter predicts the bird's next position in 3D space and updates this prediction based on new observations, thereby smoothing the trajectory and reducing noise. The state vector for the Kalman filter includes the position, velocity, and acceleration of the bird in 3D space.

In each frame, the Kalman filter predicts the 3D position of the bird. The predicted position is then compared to the new detections using Euclidean distance to find the best match for tracking. The Kalman filter is updated with the matched detection, refining the trajectory of the bird.

\begin{figure}[!ht]
  \centering
\includegraphics[width=0.5\textwidth]{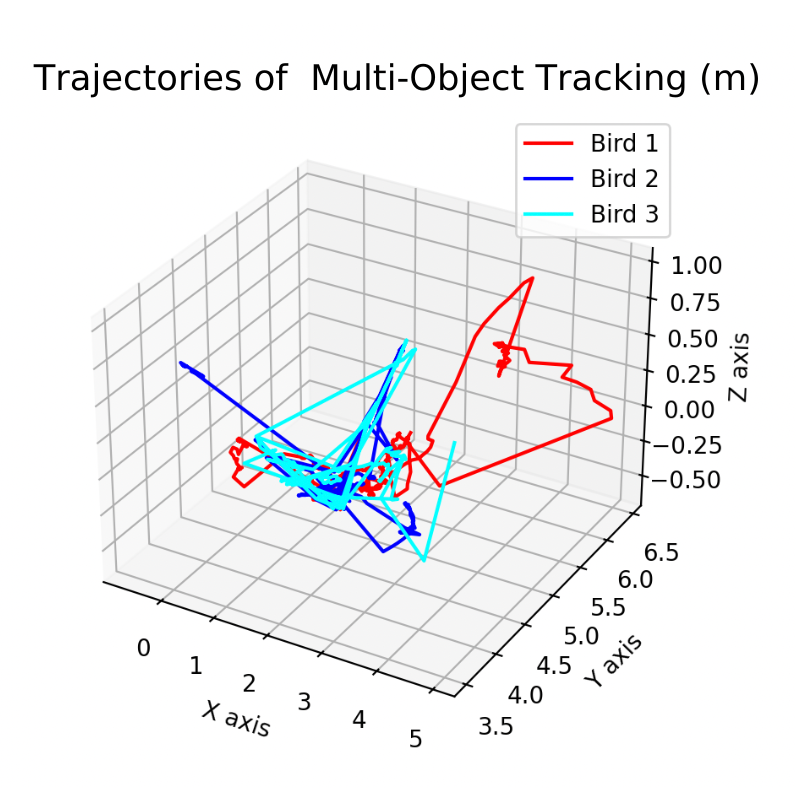}
  \caption{Graphed Trajectories of birds in Multi-view setting.}
  \label{Trajectories_Example}
\end{figure}

The integration of landmark-based matching, discussed in previous sections, plays a crucial role in the initial matching process, providing reliable correspondences between views that feed directly into the 3D triangulation and tracking pipeline. Figure \ref{Trajectories_Example} presents the final output of the tracking after the pipeline is ran completely through which shows the plots of the tracked bird, in which in the figure there were three birds that were tracked consistently over a time interval of over a minute. Note the sharp jumps indicate the identity switching, which Table \ref{tab:tracking_metrics} illustrates more depth of the experiments ran in on the Tracking accuracy. 

%% file: result.tex
\section{Experiments}
We have conducted a set of experiment to evalute the proposed pipeline:
\subsection{Keypoints Statistics and Outlier Rejection}
\begin{figure}[!ht]
  \centering
  \includegraphics[width=0.5\textwidth]{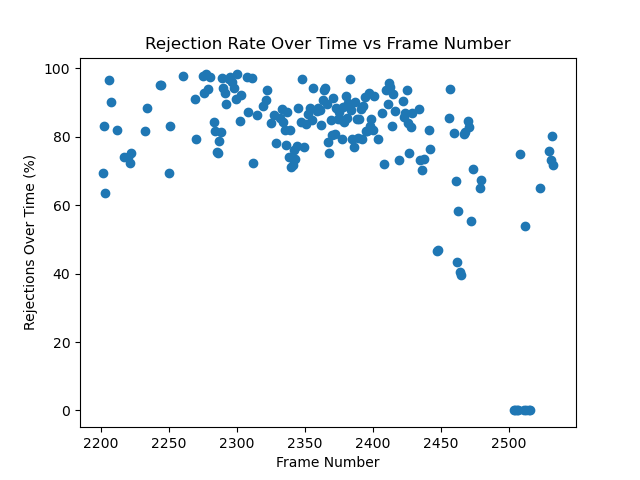}
  \caption{Outlier rejection statistics over frame interval}
  \label{fig:scatter}
\end{figure}

In our experiments, we calculate the statistics pertaining to the quantity of keypoints extracted over our interval of frames, including the minimum and maximum number of keypoints extracted, and report them in Table~\ref{tab:keypoints}. After initial keypoint extraction, we proceed with our context-aware outlier rejection step to eliminate incorrect matches. We calculate statistics pertaining to the validity of the results of this step in Table~\ref{tab:keypoints
-metrics}, and observe that the 0.97 ratio of correct final matches against all of the initial feature matches over the frame interval reflects positively on our outlier rejection methodology. Figure~\ref{fig:scatter} shows the percentage of keypoints rejected in each frame across the length of our frame interval, and offers further insight into our problem domain. As previously mentioned, one main difficulty of tracking in this environment pertains to the extreme visual similarity of the birds, meaning that the descriptors obtained for matching birds are often very similar and do not provide meaningful information to distinguish individual birds from one another. Because of this, a large percentage of incorrect keypoint matches (approx. 80\%) are expected to be removed from each frame, this corresponds with the results in Figure~\ref{fig:scatter}. 
\begin{table}[!ht]
    \centering
    \caption{GoPro3 \& GoPro5 Keypoint Statistics, frames 2200-2550} 
    \begin{tabular}{|c|c|c|}
        \hline
        Camera & GoPro3 & GoPro5 \\ \hline
        \# keypoints [min,max] & [2, 79] & [3, 81] \\ \hline
        \# keypoints (avg$\pm$std) &24$\pm$16& 40$\pm$18\\
        \hline
    \end{tabular}
    \label{tab:keypoints}
\end{table}

\begin{table}[ht!]
    \centering
    \caption{Outlier Rejection Statistics for GoPro3 \& GoPro5 pair   \label{tab:keypoints
    -metrics}} 
    \begin{tabular}{|c|c|}
        \hline
        Avg feature match rejection \% & 79.03 \\ \hline
        Std. Dev feature match rejection \% & 20.45 \\ \hline
        Ratio correct final matches / all initial matches & 0.20 \\ \hline
        Ratio correct final matches / all final matches & 0.97 \\ \hline
    \end{tabular}
  
\end{table}
\subsection{Evaluation of 3D Construction}
Furthermore, we also set up a 3D reconstruction experiment for the purpose of testing our calculations of the 3D positions of the birds.
We consider our reproduction error for this experiment, which measures how closely the 3D points re-projected on to the 2D image align with the original 2D keypoints. We collected 30 frames of data over 3 different intervals and computed their metrics in Table \ref{tab:reconstruction_metrics}. 
The average re-projection error reflects the averaged distance between the distance of the original 2D keypoints and their counterpart, which resulted in a 19.8 average reproduction error. However, we note our low standard deviation, which shows that most errors were indeed close to the average error. Additionally, we present a percentage of keypoints below a threshold of 25 px error, which was at least 54.3\% of the collected keypoint data, which further illustrates our reconstruction quality.
\subsection{Evaluation of Tracking Quality}
Finally, we evaluated the performance of the tracking step in our pipeline, which used quantitative metrics seen in Table \ref{tab:tracking_metrics} that measure the consistency of our system. We conducted our tracking experiments over 3 different intervals that had difficult situations with a high chance of an ID switch, such as multiple birds crossing over or when the birds go off screen for one camera. Our first metric was the total count of ID switches across a 1 minute video, and we see an average of 23.4 ID switches. Moreover, we counted how many birds were able to be tracked at certain time intervals, with the greatest percentages being obtained early. 
\label{sec:experiment}
\begin{table}[!ht]
\centering
\caption{Quantitative 3D Reconstruction Metrics and performance. The averaged metrics were across three intervals of 30 frames each.}
\label{tab:reconstruction_metrics}
\begin{tabular}{ l l } 
\toprule
\textbf{Metric} & \textbf{Value} \\ 
\midrule
Total Keypoints & 1,736 \\ 
Average Reprojection Error (px) & $19.8\pm 5.3$\\ 
Min Reprojection Error (px)& 9.7 \\ 
Max Reprojection Error (px)& 36.2 \\ 
\% Keypoints Below 25px Error & 54.3\% \\ 
\bottomrule
\end{tabular}
\end{table}

\begin{table}[!ht]
\centering
\caption{ Quantitative Tracking Metrics for Bird Tracking System Performance. The metrics were averaged over three intervals of the same camera video.}
\label{tab:tracking_metrics}
\begin{tabular}{ l l } 

\toprule
\textbf{Metric} & \textbf{Value} \\ 
\midrule
\textbf{Reprojection Error (pixels)} & \\ 
Mean Reprojection Error (Camera 1) & 14.32 \\ 
Mean Reprojection Error (Camera 2) & 5.03 \\ 
\midrule
\textbf{Tracking Performance} & \\ 
Total ID Switches (average) & 23.4 \\ 
Birds Tracked Over 10s (\%) & 77.1 \\ 
Birds Tracked Over 30s (\%) & 56.7 \\ 
Birds Tracked Over 60s (\%) & 26.7 \\ 
\bottomrule
\end{tabular}

\end{table}

%% file: conclusion.tex
\section{Conclusion}
\label{sec:conclusion}
We have presented a multi-view pipeline to track visually similar birds in an outdoor aviary. We proposed a novel outlier rejection method based on the environmental context using the Voronoi diagram. This Voronoi diagram is created based on landmarks in the aviary which can be defined by the user. The results show our pipeline was able to achieve a matching accuracy of $97\%$. Using this feature matching, we were able to 3D construct the bird locations accurately. Our tracking algorithm shows a promising result, confirming a robust tracking of the birds. Future work will involve testing our algorithm in different aviary datasets as well as combining it with animal behavior analyses.